\documentclass[a4paper,11pt, twocolumn]{article}
\usepackage{times}
\usepackage{latexsym}

\usepackage{url}

\usepackage{amsmath,amssymb}

\newcommand\connll{CoNLL-2003 Shared Task }

\title{Confidence penalty, annealing Gaussian noise and zoneout for biLSTM-CRF networks for named entity recognition}

\author{Antonio Jimeno Yepes \\
  IBM Research, Southgate, VIC, Australia \\
}

\date{}

\begin{document}

\maketitle
\begin{abstract}
Named entity recognition (NER) is used to identify relevant entities in text.
A bidirectional LSTM (long short term memory) encoder with a neural conditional random fields (CRF) decoder (biLSTM-CRF) is the state of the art methodology.
In this work, we have done an analysis of several methods that intend to optimize the performance of networks based on this architecture, which in some cases encourage overfitting avoidance.
These methods target exploration of parameter space, regularization of LSTMs and penalization of confident output distributions.
Results show that the optimization methods improve the performance of the biLSTM-CRF NER baseline system, setting a new state of the art performance for the \connll Spanish set with an F1 of 87.18.

\end{abstract}

\section{Introduction}

Named entity recognition (NER) identifies relevant entities of interest in text, such as people or locations.
Current state of the art results are achieved using variants of a biLSTM (bidirectional long short term memory (LSTM)) encoder and a neural linear-chain CRF (Conditional Random Field) decoder~\cite{lample2016neural} (biLSTM-CRF) based methods, which are used in a variety of tasks in natural language processing.
Even though these methods are the state of the art, methods that could help improving the performance of these networks would benefit from additional research.
Training these networks is not trivial~\cite{pascanu2013difficulty}, but there are several aspects that could be considered such as regularization methods or optimization of the system using a loss function that could approximate the target measure (e.g. F1)~\cite{tran2017named}.

We investigate recently proposed methods for the optimization of deep neural networks, and in some cases specific to LSTM (Long Short Term Memory).
These methods consider different aspects of the biLSTM-CRF systems and do not suppose a burden to the training process.

The first method consists in adding gradient noise~\cite{neelakantan2015adding} to encorage active exploration.
As second method, zoneout~\cite{krueger2016zoneout} is used, which regularizes LSTM nodes.
Finally, NER systems are typically trained to optimize the performance of a loss function, we explore penalizing confident output distributions~\cite{pereyra2017regularizing} of the loss function.

Results on the \connll set show a performance gain for English, achieving state of the art results, while a significant improvement for Spanish, setting a new state of the art for NER in Spanish using this set.

\section{Methods}

In this section, we introduce our baseline method for named entity recognition and describe the methods used to optimize its training.
Then, the data sets used in the experiment and the word embedding and language modeling are presented.

\subsection{BiLSTM-CRF NER base system}

Our system is composed of a bidirectional LSTM encoder, and a decoder that uses a neural linear-chain CRF~\cite{lample2016neural,tran2017named}.
In the encoder, we use a stack of three bidirectional LSTMs using residual connections~\cite{tran2017named}.
The input to the system includes pretrained word embeddings and word character embeddings generated by a bidirectional LSTM.

The probability of a sequence $Y_c$ over a word sequence $S$ is calculated using a softmax over all the possible sequences $\mathcal{Y}$ as shown in equation~\ref{eq:soft}.

\begin{align}
        \label{eq:soft}
        Pr(Y|S) = \frac{e^{s(Y,S)}}{\sum_{Y'\in \mathbb{Y}} e^{s(Y',S)}}
\end{align}

The log-likelihood of a predicted sequence is calculated as indicated in equation~\ref{eq:loss-bilstm-crf}.
During decoding, the Viterbi algorithm is used to identify the sequence with highest probability.

\begin{equation}
    \mathcal{L}(Y_c) = \zeta(Y_c,S) - log(\sum_{\bar{Y}\in \mathcal{Y}} e^{\zeta(\bar{Y},S)})
    \label{eq:loss-bilstm-crf}
\end{equation}



\subsection{Regularization by penalizing confident output distributions}

When training deep neural networks, algorithms may become confident of their prediction in the training set.
Penalizing confident output distributions has been proposed in classifications tasks and might be beneficial to reduce overfitting risk.

An entropy based confidence penalty derived from a classification trained model has been proposed by~\cite{pereyra2017regularizing} as shown in equation~\ref{eq:confidence-penalty}.
$x$ is the data instance, while $y_i$ is each one of the $i$ classes and $p_{\theta}$ defines the probability of class $y_i$ given $x$.

\begin{equation}
H(p_{\theta}(y|x)=-\sum_{i} p_{\theta}(y_{i}|x) \log (p_{\theta}(y_{i}|x))
\label{eq:confidence-penalty}
\end{equation}

The confidence penalty might be combined linearly with the loss function using the hyperparameter $\beta$, as shown in equation~\ref{eq:confidence-penalty-loss}, which sets the importance of the penalty.

\begin{equation}
\mathcal{L}(\theta) = - \sum \log (p_{\theta} (y|x)) - \beta H(p_{\theta}(y|x))
\label{eq:confidence-penalty-loss}
\end{equation}

In our work, there is a correct sequence $Y_c$ in word sequence $S$, but there will be a large number of incorrect ones.
The probabilities for these incorrect sequences need to be estimated, which might be costly.
We have simplified the entropy calculation and used only the correct sequence entropy and combined it to the loss $\mathcal{L}$ to generate $\mathcal{L}_{p}$, using the hyperparameter $\beta$ to control the importance of the penalty.
Penalty values $\beta=[0.1, 1.0, 2.0]$ have been evaluated in this work following~\cite{pereyra2017regularizing}.
Results show that this simplification is still effective for our problem.

\begin{equation}
    \mathcal{L}_{p}(Y_c) = \mathcal{L}(Y_c) + \beta (- P_{r}(Y_c|S) \log (P_{r}(Y_c|S)))
    \label{eq:loss-bilstm-crf-confidence}
\end{equation}

\subsection{Gradient noise}

Exploration of methods to robustly optimize neural network models is a recurrent research topic.
While there is a tradition of using noise to train classical neural networks, their impact in novel neural network architectures requires further exploration.
We research adding annealed Gaussian noise to the gradient~\cite{neelakantan2015adding}, which encourages active exploration of parameter space.
This technique is straightforward to implement in many systems and, as shown in the results, it can be effective in some cases.

The noise is added to the gradient $gt$ as indicated in equation~\ref{eq:gradient-noise}, the scheduled annealed Gaussian noise is inspired by~\cite{welling2011bayesian}.

\begin{equation}
g_{t} \leftarrow g_{t} + N(0,\sigma_{t}^{2})
\label{eq:gradient-noise}
\end{equation}

The noise is normal with zero mean and standard deviation estimated as indicated in equation~\ref{eq:noise-std}.
The $\eta$ parameter controls the value of the noise and it has been set to $[$0.01, 0.3 and 1.0$]$ and $\gamma = 0.55$~\cite{neelakantan2015adding}.

\begin{equation}
\sigma_{t}^{2}=\frac{\eta}{(1+t)^{\gamma}}
\label{eq:noise-std}
\end{equation}

Higher gradient noise at the beginning forces the gradient away of 0 in early stages.
The noise decreases overtime controlled by parameter $t$.

\subsection{Zoneout}

We have considered zoneout~\cite{krueger2016zoneout} in LSTM, which uses random noise to train a pseudo-ensemble in recurrent neural networks.

In LSTM~\cite{hochreiter1997long}, at each timestep $t$, the hidden state is divided into a memory vector $c_t$ and a hidden vector $h_t$.
Formulation of LSTMs contains the implementation of a set of gates that control the flow of information.
These gates include an input gate $i_t$, an output gate $o_t$ and a forget gate $f_t$ over the previous hidden units and data entry $x_{t}$.
A set of weight matrices and bias terms are learnt during training.

\begin{eqnarray}
i_{t},f_{t},o_{t}=\sigma(W_{x}x_{t}+W_{h}h_{t-1} +b) \\
g_{t}=tanh(W_{xg}x_t+W_{hg}h_{t-1}+b_{g}) \\
c_{t}=f_{t} \odot c_{t-1} + i_{t} \odot g_{t} \\
h_{t} = o_{t} \odot tanh (c_{t})
\end{eqnarray}

Zoneout connects the previous time step information from $c_{t-1}$ and $h_{t-1}$ with the current $c_t$ and $h_t$. 
$d^{c}_{t}$ and $d^{h}_{t}$, as shown below, are masks generated at each timestep $t$ using a binomial distribution with $n=1$ and $p$ with values between $[0,1]$ for $c_t$ and $h_t$ respectively.

$zc$ and $zh$ are the parameters used to define the probability for mask generation for $c_{t}$ and $h_{t}$.
Values used in our work are $zc=zh=0.15$ and $zc=0.5,zh=0.05$ as used in~\cite{krueger2016zoneout}.
Equations~\ref{eq:zoneout-begin}-\ref{eq:zoneout-end} show the implementation of zoneout for LSTM.
It uses random noise to train a pseudo-ensemble, as in dropout, 
but as it keeps hidden units, gradient information and state information are more readily propagated through time, as
in feedforward stochastic deep networks.

\begin{eqnarray}
\label{eq:zoneout-begin}
\hat{c}_{t} = (f_{t} \odot c_{t-1} + i_{t} \odot g{t}) \\
c_{t} = d^{c}_{t} \odot c_{t-1} + (1-d^{c}_{t}) \odot  \hat{c}_{t} \\
\hat{h}_{t} = (o_{t} \odot tanh (f_{t} \odot c_{t-1} + i_{t} \odot g_{t})) \\
h_{t} = d^{h}_{t} \odot h_{t-1} + (1-d^{h}_{t}) \odot \hat{h}_{t}
\label{eq:zoneout-end}
\end{eqnarray}

\subsection{Data sets}

We have prepared and evaluated the proposed methods on the English and Spanish sets of the \connll Named Entity Recognition set\mbox{~\cite{tjong2003introduction}}\footnote{http://www.cnts.ua.ac.be/conll2003/ner}.
We have followed the training, development and test set configuration of \connll set.
The Spanish dataset has 8323/1915/1517 sentences in train/dev/test sets respectively. The English dataset is almost twice as large with 14041/3250/3453 sentences in train/dev/test set. For all of our models, the word-embedding size is set to 100 for English and 64 for Spanish.

\subsection{Word embedding}

English word embedding was obtained from Word2vec-api\footnote{https://github.com/3Top/word2vec-api/blob/master/README.md}. The embedding dimension is 100 and it was trained using GloVe with AdaGrad.

For the generation of Spanish word embeddings we followed~\cite{lample2016neural}, using
Spanish Gigaword Third Edition\footnote{https://catalog.ldc.upenn.edu/ldc2011t12} as corpus with an embedding dimension of 64, a minimum word frequency cutoff of 4 and a window size of 8.

\subsection{Language Modeling}

We have used both forward and backward language models (LM) as additional input for our system.
Language models have been successfully used in similar tasks previously~\cite{peters2017semi,tran2017named}.
The English forward language model was obtained from\footnote{https://github.com/tensorflow/models/tree/master/lm\_1b} using the One billion word benchmark\footnote{https://github.com/ciprian-chelba/1-billion-word-language-modeling-benchmark}~\cite{chelba2013one} and has a perplexity of 30.
The backward English language model and the Spanish forward and backward ones were generated using an LSTM based baseline\footnote{https://github.com/rafaljozefowicz/lm}~\cite{jozefowicz2016exploring}.
This code estimates a forward language model and was adapted to estimate as well a backward language model. Language models were estimated using the One billion word benchmark.
The vocabulary for the backward English model is the same as the pregenerated forward model.
The perplexity for the estimated backward English language model is 46.
The vocabulary for the Spanish language models has been generated using tokens with frequency $>$ 2.
The perplexity for the forward and backward Spanish language models are 56 and 57 respectively.

\section{Results}

We present results on both English and Spanish sets using the \connll NER set.
The training set has been used to train the system using several hyperparameter configurations, the development set has been used to select the best configuration and the reported performance of the final system is based on the test set.

For all of our models, the word-embedding size is set to 100 for English and 64 for Spanish. The hidden vector size is 100 for both English and Spanish sets without the LM embeddings.
With the LM embeddings, the hidden vector size is changed to 300. 
We trained the model with Stochastic Gradient Descent with momentum, using the learning rate of 0.005.
Statistical significance has been determined using a randomization version of the paired sample t-test~\cite{cohen1996empirical}.

F1 results are shown in table~\ref{tab:result-conll2003}.
The baseline system is the biLSTM-CRF method.
Penalty improves significantly the baseline performance when $\beta = 1$ for both English and Spanish sets.
Adding noise to the gradients has a non-significant improvement for English, except when $\eta = 1.0$.
A similar performance increase is observed in Spanish.
Zoneout significantly improves results on the Spanish set, performance increases are not significant for the English Set.
When combining penalty $\beta=1.0$, noise $\eta=0.01$ and zoneout $zc=zh=0.15$ the increase in performance for Spanish is quite significant, setting a new state of the art result for the Spanish set with an F1 of 87.18.

\begin{table}
\small
\centering
\begin{tabular}{|l|r|r|}
\hline \bf System                      & \bf English & \bf Spanish \\ \hline
Baseline system                     & 91.03       & 86.16 \\
\hline
Penalty $\beta=0.1$                    & 90.93       & 86.27 \\
Penalty $\beta=1.0$               & 91.19       & 86.47 \\
Penalty $\beta=2.0$              & 91.05       & 86.08 \\
\hline
Noise $\eta=0.01$                 & 91.06       & 86.31 \\
Noise $\eta=0.3$                   & 91.02       & 86.32 \\ 
Noise $\eta=1.0$                   & 90.79       & 85.99 \\  
\hline
Zoneout $zc=zh=0.15$        & 90.98       & 86.49 \\
Zoneout $zc=0.5; zh=0.05$& 90.89       & 86.42 \\
\hline
COMBINED                             & \bf{91.22}       & {\bf 87.18} \\
\hline
\hline
Baseline+LM                           & 91.66      & 85.83        \\
\hline
COMBINED+LM                      & {\bf 91.96}     & {\bf 86.56} \\
\hline
\end{tabular}
\caption{F1 results on Spanish \connll NER test set for English and Spanish.
COMBINED stands for (Penalty $\beta=1.0$ + Noise $\eta=0.01$ + Zoneout $zc=zh=0.15$).
COMBINED+LM stands for COMBINED configuration and language models.}
\label{tab:result-conll2003}
\end{table}

\section{Discussion}

Overall, the proposed methods improve over the baseline system.
The combination of the proposed methods in the \connll set for Spanish sets a new state of the art result that significantly improves over previous results.


Adding a penalty to the loss function seems to be the most relevant method for improving on the English set, which seems to improve as well the performance on the Spanish set.
Adding noise has not such a strong impact but it is still able of providing an improvement of both sets.
Zoneout has the strongest performance improvement on the Spanish set, even though the improvement on the English set is not that significant.

For the English set, the performance improves but in most cases is not as significant as with the results obtained with the Spanish set.
The training set for Spanish is smaller and this could explain the improved performance by the proposed methods.

There are some configurations in which the results do not significantly change respect to the baseline result.
Examples are when the level of noise ($\eta=1.0$) or penalty ($\beta=2.0$) are high enough to prevent finding a better trained configuration of the model.

Using forward and backward language models improve the performance on the English set but decreases the performance of the Spanish set, as seen in previous work~\cite{tran2017named}.
Using the modifications proposed in this work, both results for English and Spanish using language models improve, again the improvements are more significant for the Spanish set.
Compared to previous work, the best performance with the English set was obtained by~\cite{peters2017semi} with an F1 of 91.93, comparable to our result with an F1 of 91.96.

On the Spanish set, the previous state of the art result was obtained with the biLSTM CRF system with residual connections and trainable bias decoding with an F1 of 86.31~\cite{tran2017named}.
The modifications presented in our work improve this result by a significant margin, with a performance of 87.18, setting a new state of the art result.

\section{Conclusions and Future Work}

We have presented a set of methods that help improving the training of biLSTM-CRF systems applied to named entity recognition.
Our initial investigation shows that these methods improve the baseline system on the \connll set and in the case of the Spanish set, provides a new state of the art result with an F1 of 87.18.

The optimization methods presented in this work are not specific to named entity recognition and they might be applied to similar network architectures for different tasks~\cite{Huang2015BidirectionalLM} or more complex networks for named entity recognition~\cite{liu2017capturing}.
Additional regularization and optimization methods could be considered as shown in~\cite{merity2017regularizing}.

These networks typically optimize a loss function but they are evaluated using a different measure, such as F1.
Previous work~\cite{tran2017named} has tried to use a bias in decoding after training the system. We would like to explore ways into which the target evaluation measure might be better integrated in the training.

\bibliographystyle{plain}
\bibliography{bibliography}

\begin{thebibliography}{10}

\bibitem{tran2017named}
Authors-reference.
\newblock Authors reference.
\newblock {\em Authors reference}, 2017.

\bibitem{chelba2013one}
Ciprian Chelba, Tomas Mikolov, Mike Schuster, Qi~Ge, Thorsten Brants, Phillipp
  Koehn, and Tony Robinson.
\newblock One billion word benchmark for measuring progress in statistical
  language modeling.
\newblock {\em arXiv preprint arXiv:1312.3005}, 2013.

\bibitem{cohen1996empirical}
Paul~R Cohen.
\newblock Empirical methods for artificial intelligence.
\newblock {\em IEEE Intelligent Systems}, (6):88, 1996.

\bibitem{hochreiter1997long}
Sepp Hochreiter and J{\"u}rgen Schmidhuber.
\newblock Long short-term memory.
\newblock {\em Neural computation}, 9(8):1735--1780, 1997.

\bibitem{Huang2015BidirectionalLM}
Zhiheng Huang, Wei Xu, and Kai Yu.
\newblock Bidirectional lstm-crf models for sequence tagging.
\newblock {\em CoRR}, abs/1508.01991, 2015.

\bibitem{jozefowicz2016exploring}
Rafal Jozefowicz, Oriol Vinyals, Mike Schuster, Noam Shazeer, and Yonghui Wu.
\newblock Exploring the limits of language modeling.
\newblock {\em arXiv preprint arXiv:1602.02410}, 2016.

\bibitem{krueger2016zoneout}
David Krueger, Tegan Maharaj, J{\'a}nos Kram{\'a}r, Mohammad Pezeshki, Nicolas
  Ballas, Nan~Rosemary Ke, Anirudh Goyal, Yoshua Bengio, Hugo Larochelle, Aaron
  Courville, et~al.
\newblock Zoneout: Regularizing rnns by randomly preserving hidden activations.
\newblock {\em arXiv preprint arXiv:1606.01305}, 2016.

\bibitem{lample2016neural}
Guillaume Lample, Miguel Ballesteros, Sandeep Subramanian, Kazuya Kawakami, and
  Chris Dyer.
\newblock Neural architectures for named entity recognition.
\newblock In {\em Proceedings of NAACL-HLT}, pages 260--270, 2016.

\bibitem{liu2017capturing}
Fei Liu, Timothy Baldwin, and Trevor Cohn.
\newblock Capturing long-range contextual dependencies with memory-enhanced
  conditional random fields.
\newblock In {\em Proceedings of the Eighth International Joint Conference on
  Natural Language Processing (Volume 1: Long Papers)}, pages 555--565. Asian
  Federation of Natural Language Processing, 2017.

\bibitem{merity2017regularizing}
Stephen Merity, Nitish~Shirish Keskar, and Richard Socher.
\newblock Regularizing and optimizing lstm language models.
\newblock {\em arXiv preprint arXiv:1708.02182}, 2017.

\bibitem{neelakantan2015adding}
Arvind Neelakantan, Luke Vilnis, Quoc~V Le, Ilya Sutskever, Lukasz Kaiser,
  Karol Kurach, and James Martens.
\newblock Adding gradient noise improves learning for very deep networks.
\newblock {\em arXiv preprint arXiv:1511.06807}, 2015.

\bibitem{pascanu2013difficulty}
Razvan Pascanu, Tomas Mikolov, and Yoshua Bengio.
\newblock On the difficulty of training recurrent neural networks.
\newblock In {\em International Conference on Machine Learning}, pages
  1310--1318, 2013.

\bibitem{pereyra2017regularizing}
Gabriel Pereyra, George Tucker, Jan Chorowski, {\L}ukasz Kaiser, and Geoffrey
  Hinton.
\newblock Regularizing neural networks by penalizing confident output
  distributions.
\newblock {\em arXiv preprint arXiv:1701.06548}, 2017.

\bibitem{peters2017semi}
Matthew~E Peters, Waleed Ammar, Chandra Bhagavatula, and Russell Power.
\newblock Semi-supervised sequence tagging with bidirectional language models.
\newblock {\em arXiv preprint arXiv:1705.00108}, 2017.

\bibitem{tjong2003introduction}
Erik~F Tjong Kim~Sang and Fien De~Meulder.
\newblock Introduction to the conll-2003 shared task: Language-independent
  named entity recognition.
\newblock In {\em Proceedings of the seventh conference on Natural language
  learning at HLT-NAACL 2003-Volume 4}, pages 142--147. Association for
  Computational Linguistics, 2003.

\bibitem{welling2011bayesian}
Max Welling and Yee~W Teh.
\newblock Bayesian learning via stochastic gradient langevin dynamics.
\newblock In {\em Proceedings of the 28th International Conference on Machine
  Learning (ICML-11)}, pages 681--688, 2011.

\end{thebibliography}

\end{document}